\documentclass[11pt,a4paper]{article}
\usepackage[hyperref]{acl2020}
\usepackage{times}
\usepackage{latexsym}

\usepackage{microtype}
\usepackage{multirow}
\usepackage{graphicx}
\usepackage{amsmath}
\usepackage{amsfonts}
\usepackage{amssymb}
\usepackage{textcomp}
\usepackage{xcolor}

\usepackage{algorithm}
\usepackage{algorithmic}

\aclfinalcopy 
\def\aclpaperid{1377} 

\title{AMR Parsing via Graph{\small $\leftrightarrows$}Sequence Iterative Inference\thanks{~The work described in this paper is substantially supported by grants from the Research Grant Council of the Hong Kong Special Administrative Region, China (Project Code: 14204418) and the Direct Grant of the Faculty of Engineering, CUHK (Project Code: 4055093).}}

\author{Deng Cai \\
	The Chinese University of Hong Kong\\
	{\tt thisisjcykcd@gmail.com} \\\And
	Wai Lam \\
	The Chinese University of Hong Kong\\
	{\tt wlam@se.cuhk.edu.hk} \\}

\date{}

\begin{document}
	\maketitle
	\begin{abstract}
		We propose a new end-to-end model that treats AMR parsing as a series of dual decisions on the input sequence and the incrementally constructed graph. At each time step, our model performs multiple rounds of attention, reasoning, and composition that aim to answer two critical questions: (1) which part of the input \textit{sequence} to abstract; and (2) where in the output \textit{graph} to construct the new concept. We show that the answers to these two questions are mutually causalities. We design a model based on iterative inference that helps achieve better answers in both perspectives, leading to greatly improved parsing accuracy. Our experimental results significantly outperform all previously reported \textsc{Smatch} scores by large margins. Remarkably, without the help of any large-scale pre-trained language model (e.g., BERT), our model already surpasses previous state-of-the-art using BERT. With the help of BERT, we can push the state-of-the-art results to 80.2\% on LDC2017T10 (AMR 2.0) and 75.4\% on LDC2014T12 (AMR 1.0).
	\end{abstract}
	\section{Introduction}
\begin{figure}[t]
		\centering
		\includegraphics[scale=0.35]{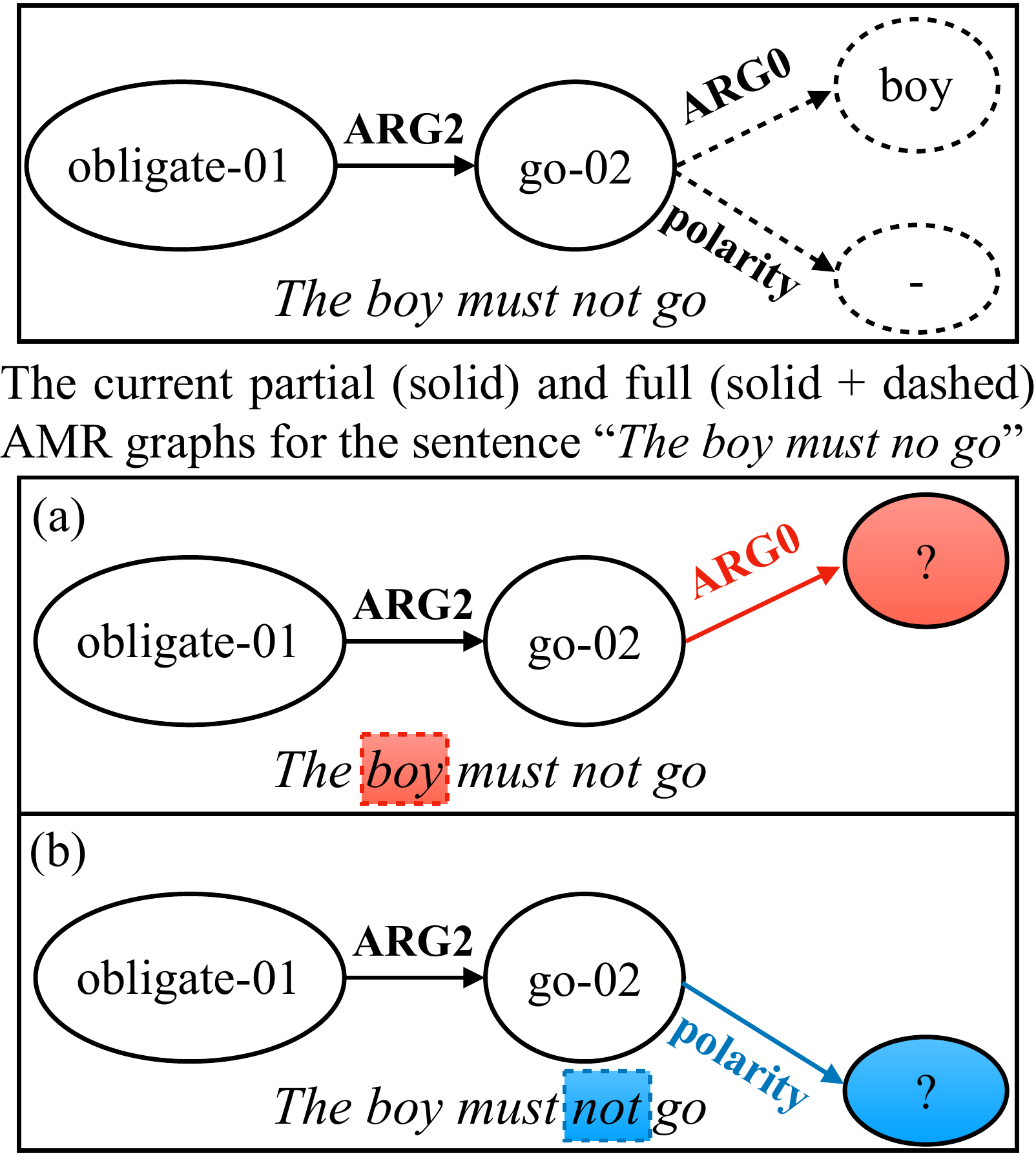}
		\caption{AMR graph construction given the partially constructed graph: (a) one possible expansion resulting in the \texttt{boy} concept. (b) another possible expansion resulting in the \texttt{-} (negation) concept.}
		\label{example}
	\end{figure}

	Abstract Meaning Representation (AMR) \cite{banarescu2013abstract} is a broad-coverage semantic formalism that encodes the meaning of a sentence as a rooted, directed, and labeled graph, where nodes represent concepts and edges represent relations (See an example in Figure \ref{example}). AMR parsing is the task of transforming natural language text into AMR. One biggest challenge of AMR parsing is the lack of explicit alignments between nodes (concepts) in the graph and words in the text. This characteristic not only poses great difficulty in concept prediction but also brings a close tie for concept prediction and relation prediction.
	
	While most previous works rely on a pre-trained aligner to train a parser, some recent attempts include: modeling the alignments as latent variables \cite{lyu2018amr}, attention-based sequence-to-sequence transduction models \cite{barzdins2016riga,konstas2017neural,van2017neural}, and attention-based sequence-to-graph transduction models \cite{cai-lam-2019-core,zhang-etal-2019-broad}. Sequence-to-graph transduction models build a semantic graph incrementally via spanning one node at every step. This property is appealing in terms of both computational efficiency and cognitive modeling since it mimics what human experts usually do, i.e., first grasping the core ideas then digging into more details \cite{banarescu2013abstract,cai-lam-2019-core}.
	
	Unfortunately, the parsing accuracy of existing works including recent state-of-the-arts \cite{zhang-etal-2019-amr,zhang-etal-2019-broad} remain unsatisfactory compared to human-level performance,\footnote{The average annotator vs. inter-annotator agreement (\textsc{Smatch}) was 0.83 for newswire and 0.79 for web text according to \newcite{banarescu2013abstract}.}  especially in cases where the sentences are rather long and informative, which indicates substantial room for improvement. One possible reason for the deficiency is the inherent defect of one-pass prediction process; that is, the lack of the modeling capability of the interactions between concept prediction and relation prediction, which is critical to achieving fully-informed and unambiguous decisions.
	
	We introduce a new approach tackling AMR parsing, following the incremental sequence-to-graph transduction paradigm. We explicitly characterize each spanning step as the efforts for finding \textit{which part to abstract with respect to the input sequence}, and \textit{where to construct with respect to the partially constructed output graph}. Equivalently, we treat AMR parsing as a series of dual decisions on the input sequence and the incrementally constructed graph. Intuitively, the answer of what concept to abstract decides where to construct (i.e., the relations to existing concepts), while the answer of where to construct determines what concept to abstract. Our proposed model, supported by neural networks with explicit structure for attention, reasoning, and composition, integrated with an iterative inference algorithm. It iterates between finding supporting text pieces and reading the partially constructed semantic graph, inferring more accurate and harmonious expansion decisions progressively. Our model is aligner-free and can be effectively trained with limited amount of labeled data. Experiments on two AMR benchmarks demonstrate that our parser outperforms the previous best parsers on both benchmarks. It achieves the best-reported \textsc{Smatch} scores (F1): 80.2\% on LDC2017T10 and 75.4\% on LDC2014T12, surpassing the previous state-of-the-art models by large margins.
	\section{Related Work \& Background}
	On a coarse-grained level, we can categorize existing AMR parsing approaches into two main classes: Two-stage parsing \cite{flanigan2014discriminative, lyu2018amr,zhang-etal-2019-amr} uses a pipeline design for concept identification and relation prediction, where the concept decisions precede all relation decisions; One-stage parsing constructs a parse graph incrementally. For more fine-grained analysis, those one-stage parsing methods can be further categorized into three types: Transition-based parsing \cite{wang2016camr,damonte2016incremental,ballesteros2017amr,peng-etal-2017-addressing,guo2018better,liu2018amr,wang2017getting,naseem-etal-2019-rewarding} processes a sentence from left-to-right and constructs the graph incrementally by alternately inserting a new node or building a new edge. Seq2seq-based parsing \cite{barzdins2016riga,konstas2017neural,van2017neural,peng-etal-2018-sequence} views parsing as sequence-to-sequence transduction by some linearization of the AMR graph. The concept and relation prediction are then treated equally with a shared vocabulary. The third class is graph-based parsing \cite{cai-lam-2019-core,zhang-etal-2019-broad}, where at each time step, a new node along with its connections to existing nodes are jointly decided, either in order \cite{cai-lam-2019-core} or in parallel \cite{zhang-etal-2019-broad}. So far, the reciprocal causation of relation prediction and concept prediction has not been closely-studied and well-utilized.
	
	There are also some exceptions staying beyond the above categorization. \newcite{peng2015synchronous}  introduce a synchronous hyperedge replacement grammar solution. \newcite{pust2015parsing} regard the task as a machine translation problem, while \newcite{artzi2015broad} adapt combinatory categorical grammar. \newcite{groschwitz2018amr,lindemann-etal-2019-compositional} view AMR graphs as the structure AM algebra.
	\begin{figure*}[t]
		\centering
		\includegraphics[scale=0.43]{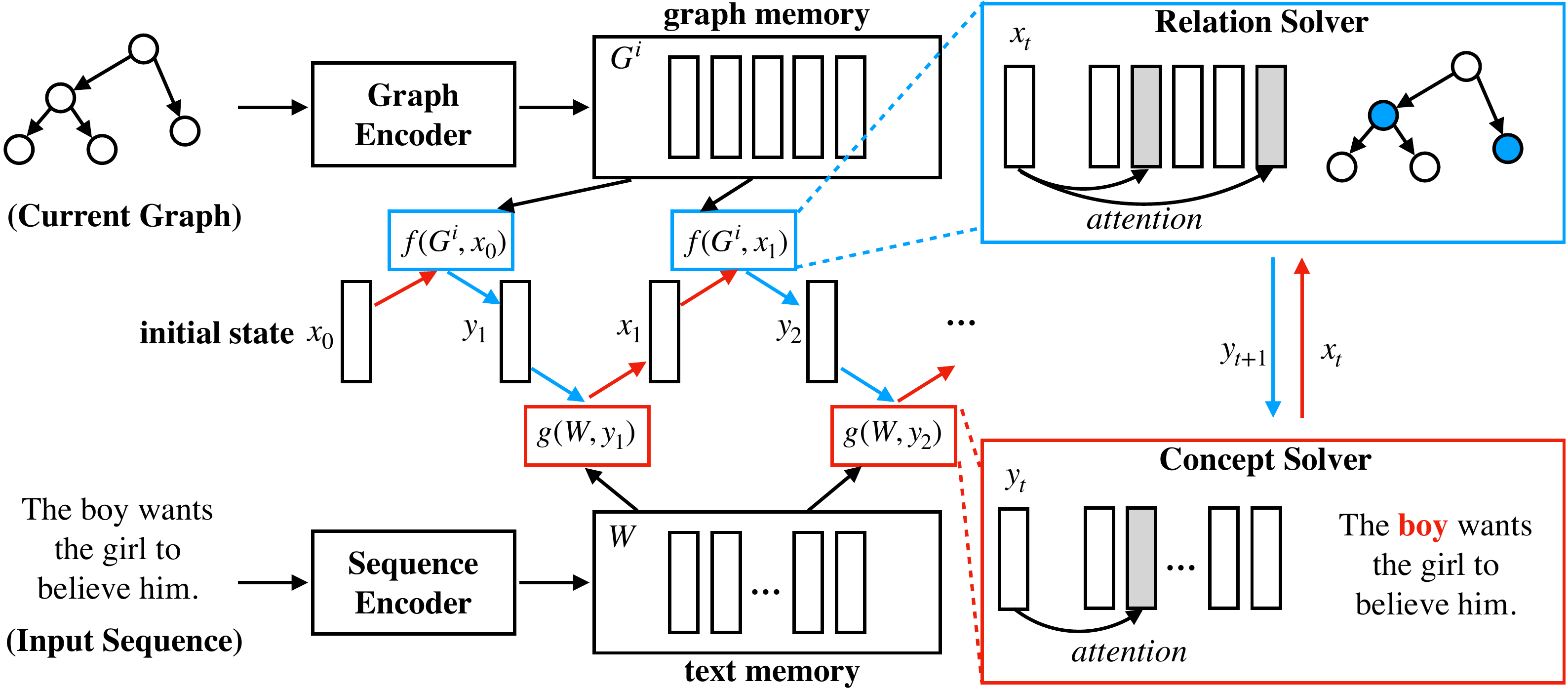}
		\caption{Overview of the dual graph-sequence iterative inference for AMR parsing. Given the current graph $G^i$ and input sequence $W$. The inference starts with an initial concept decision $x_0$ and follows the inference chain $x_0 \rightarrow f(G^i, x_0) \rightarrow y_1 \rightarrow g(W, y_1) \rightarrow x_1 \rightarrow f(G^i, x_1) \rightarrow y_2 \rightarrow g(W, y_2) \rightarrow \cdots$. The details of $f$ and $g$ are shown in red and blue boxes, where nodes in graph and tokens in sequence are selected via attention mechanisms.}
		\label{arch}
	\end{figure*}
	\section{Motivation}
	Our approach is inspired by the deliberation process when a human expert is deducing a semantic graph from a sentence. The output graph starts from an empty graph and spans incrementally in a node-by-node manner. At any time step of this process, we are distilling the information for the next expansion. We call it expansion because the new node, as an abstract concept of some specific text fragments in the input sentence, is derived to complete some missing elements in the current semantic graph. Specifically, given the input sentence and the current partially constructed graph, we are answering two critical questions: which part of the input \textit{sequence} to abstract, and where in the output \textit{graph} to construct the new concept. For instance, Figure \ref{example}(a) and (b) show two possible choices for the next expansion. In Figure \ref{example}(a), the word ``boy" is abstracted to the concept \texttt{boy} to complement the subject information of the event \texttt{go-02}. On the other hand, in Figure \ref{example}(b), a polarity attribute of the event \texttt{go-2} is constructed, which is triggered by the word ``not" in the sentence.
	
	We note that the answer to one of the questions can help answer the other. For instance,  if we have decided to render the word ``not" to the graph, then we will consider adding an edge labeled as \texttt{polarity}, and finally determine its attachment to the existing event \texttt{go-2} (rather than an edge labeled \texttt{ARG0} to the same event \texttt{go-2}, though it is also present in the golden graph). On the other hand, if we have decided to find the subject (\texttt{ARG0}  relation) of the action \texttt{go-02}, we are confident to locate the word ``boy" instead of function words like ``not" or ``must", thus unambiguously predict the right concept \texttt{boy}. Another possible circumstance is that we may make a mistake trying to ask something that is not present in the sentence (e.g., the destination of the \texttt{go-02} action). This attempt will be rejected by a review of the sentence. The rationale is that literally we cannot find the destination information in the sentence. Similarly, if we mistakenly propose to abstract some parts of the sentence that are not ready for construction yet, the proposal will be rejected by another inspection on the graph since that there is nowhere to place such a new concept.
	
	We believe the mutual causalities, as described above, are useful for action disambiguation and harmonious decision making, which eventually result in more accurate parses. We formulate AMR parsing as a series of dual graph-sequence decisions and design an iterative inference approach to tackle each of them. It is sort of analogous to the cognition procedure of a person, who might first notice part of the important information in one side (graph or sequence), then try to confirm her decision at the other side, which could just refute her former hypothesis and propose a new one, and finally converge to a conclusion after multiple rounds of reasoning.
	\section{Proposed Model}
	\subsection{Overview}
	Formally, the parsing model consists of a series of graph expansion procedures $\{G^0\rightarrow \ldots \rightarrow G^i \rightarrow\ldots\}$, starting from an empty graph $G^0$. In each turn of expansion, the following iterative inference process is performed:
	\begin{align*}
	y^{i}_{t} &= g(G^{i}, x^{i}_{t}), \\
	x^{i}_{t+1} &= f(W, y^{i}_t),
	\end{align*}
	where $W, G^{i}$ are the input sequence and the current semantic graph respectively. $g(\cdot), f(\cdot)$ seek where to construct (edge prediction) and what to abstract (node prediction) respectively, and $x^{i}_t, y^{i}_t$ are the $t$-th graph hypothesis (where to construct) and $t$-th sequence hypothesis (what to abstract) for the $i$-th expansion step respectively. For clarity, we may drop the superscript $i$ in the following descriptions.
	
	Figure \ref{arch} depicts an overview of the graph-sequence iterative inference process. Our model has four main components:  (1) Sequence Encoder, which generates a set of text memories (per token) to provide grounding for concept alignment and abstraction; (2) Graph Encoder, which generates a set of graph memories (per node) to provide grounding for relation reasoning; (3) Concept Solver, where a previous graph hypothesis is used for concept prediction; and (4) Graph Solver, where a previous concept hypothesis is used for relation prediction. The last two components correspond to the reasoning functions $g(\cdot)$ and $f(\cdot)$ respectively.
	
	The text memories can be computed by Sentence Encoder at the beginning of the whole parsing while the graph memories are constructed by Graph Encoder incrementally as the parsing progresses. During the iterative inference, a semantic representation of current state is used to attend to both graph and text memories (blue and red arrows) in order to locate the new concept and obtain its relations to the existing graph, both of which subsequently refine each other. Intuitively, after a first glimpse of the input sentence and the current graph, specific sub-areas of both sequence and graph are revisited to obtain a better understanding of the current situation. Later steps typically read the text in detail with specific learning aims, either confirming or overturning a previous hypothesis. Finally, after several iterations of reasoning steps, the refined sequence/graph decisions are used for graph expansion.
	\subsection{Sequence Encoder}
	As mentioned above, we employ a sequence encoder to convert the input sentence into vector representations. The sequence encoder follows the multi-layer Transformer architecture described in \newcite{vaswani2017attention}. At the bottom layer, each token is firstly transformed into the concatenation of features learned by a character-level convolutional neural network \cite[charCNN,][]{kim2016character} and randomly initialized embeddings for its lemma, part-of-speech tag, and named entity tag. Additionally, we also include features learned by pre-trained language model BERT \cite{devlin-etal-2019-bert}.\footnote{We obtain word-level representations from pre-trained BERT in the same way as \newcite{zhang-etal-2019-amr,zhang-etal-2019-broad}, where sub-token representations at the last layer are averaged.}
	
	Formally, for an input sequence $w_1, w_2, \ldots, w_n$ with length $n$, we insert a special token \texttt{BOS} at the beginning of the sequence. For clarity, we omit the detailed transformations \cite{vaswani2017attention} and denote the final output from our sequence encoder as $\{h_0, h_1, \ldots, h_n\} \in\mathbb{R}^d$, where $h_0$ corresponds the special token \texttt{BOS} and serves as an overall representation while others are considered as contextualized word representations. Note that the sequence encoder only needs to be invoked once, and the produced text memories are used for the whole parsing procedure.
	\subsection{Graph Encoder}
	We use a similar idea in \newcite{cai-lam-2019-core} to encode the incrementally expanding graph. Specifically, a graph is simply treated as a sequence of nodes (concepts) in the chronological order of when they are inserted into the graph. We employ multi-layer Transformer architecture with masked self-attention and source-attention, which only allows each position in the node sequence to attend to all positions up to and including that position, and every position in the node sequence to attend over all positions in the input sequence.\footnote{It is analogous to a standard Transformer decoder \cite{vaswani2017attention} for sequence-to-sequence learning.} While this design allows for significantly more parallelization during training and computation-saving incrementality during testing,\footnote{Trivially employing a graph neural network here can be computationally expensive and intractable since it needs to re-compute all graph representations after every expansion.} it inherently neglects the edge information. We attempted to alleviate this problem by incorporating the idea of \newcite{strubell-etal-2018-linguistically} that applies auxiliary supervision at attention heads to encourage them to attend to each node’s parents in the AMR graph. However, we did not see performance improvement. We attribute the failure to the fact that the neural attention mechanisms on their own are already capable of learning to attend to useful graph elements, and the auxiliary supervision is likely to disturb the ultimate parsing goal.
	
	Consequently, for the current graph $G$ with $m$ nodes, we take its output concept sequence $c_1, c_2, \ldots, c_m$ as input. Similar to the sequence encoder, we insert a special token \texttt{BOG} at the beginning of the concept sequence. Each concept is firstly transformed into the concatenation of feature vector learned by a char-CNN and randomly initialized  embedding. Then, a multi-layer Transformer encoder with masked self-attention and source-attention is applied, resulting in vector representations $\{s_0, s_1, \ldots, s_m\}\in\mathbb{R}^d$, where $s_0$ represents the special concept \texttt{BOG} and serves as a dummy node while others are considered as contextualized node representations.
	\subsection{Concept Solver}
	At each sequence reasoning step $t$, the concept solver receives a state vector $y_t$ that carries the latest graph decision and the input sequence memories $h_1, \ldots, h_n$ from the sequence encoder, and aims to locate the proper parts in the input sequence to abstract and generate a new concept. We employ the scaled dot-product attention proposed in \newcite{vaswani2017attention} to solve this problem. Concretely, we first calculate an attention distribution over all input tokens:
	\begin{equation}
	\alpha_t = \text{softmax}(\frac{(W^{Q}y_t)^{\text{T}} W^{K} h_{1:n}}{\sqrt{d_k}}),
	\nonumber
	\end{equation}
	where $\{W^{Q}, W^{K}\}\in \mathbb{R}^{d_k \times d}$ denote learnable linear projections that transform the input vectors into the query and key subspace respectively, and $d_k$ represents the dimensionality of the subspace.
	
	The attention weights $\alpha_t\in \mathbb{R}^n$ provide a soft alignment between the new concept and the tokens in the input sequence. We then compute the probability distribution of the new concept label through a hybrid of three channels.
	First, $\alpha_t$ is fed through an MLP and softmax to obtain a probability distribution over a pre-defined vocabulary:
	\begin{gather}
	\text{MLP}(\alpha_t) = (W^{V}h_{1:n}) \alpha_t  + y_t  \label{mlp} \\
	P^{\text{(vocab)}} = \text{softmax}( W^{\text{(vocab)}}\text{MLP}(\alpha_t) + b^{\text{(vocab)}}), \nonumber
	\end{gather}
	where $W^{V}\in \mathbb{R}^{d \times d}$ denotes the learnable linear projection that transforms the text memories into the value subspace, and the value vectors are averaged according to $\alpha_t$ for concept label prediction. Second, the attention weights $\alpha_t$ directly serve as a copy mechanism \cite{gu2016incorporating,see-etal-2017-get}, i,e., the probabilities of copying a token lemma from the input text as a node label. Third, to address the attribute values such as person names or numerical strings, we also use $\alpha_t$ for another copy mechanism that directly copies the original strings of input tokens. The above three channels are combined via a soft switch to control the production of the concept label from different sources:
	\begin{equation}
	[p_{0}, p_{1}, p_{2}] = \text{softmax}( W^{\text{(switch)}}\text{MLP}(\alpha_t)),
	\nonumber
	\end{equation}
	where \text{MLP} is the same as in Eq. \ref{mlp}, and $p_{0}, p_{1}$ and $p_{2}$ are the probabilities of three prediction channels respectively.
	Hence, the final prediction probability of a concept $c$ is given by:
	\begin{align*}
	P(c) = &p_{0} \cdot P^{\text{(vocab)}}(c) \\ 
	+ & p_1  \cdot (\sum_{i \in L(c)}\alpha_t[i]) + p_2 \cdot(\sum_{i \in T(c) }\alpha_t[i]),
	\end{align*} 
	where $[i]$ indexes the $i$-th element and $L(c)$ and $T(c)$ are index sets of lemmas and tokens respectively that have the surface
	form as $c$.
	\subsection{Relation Solver}
	At each graph reasoning step $t$, the relation solver receives a state vector $x_t$ that carries the latest concept decision and the output graph memories $s_0, s_1, \ldots, s_m$ from the graph encoder, and aims to point out the nodes in the current graph that have an immediate relation to the new concept (source nodes) and generate corresponding edges. Similar to \newcite{cai-lam-2019-core,zhang-etal-2019-broad}, we factorize the task as two stages: First, a relation identification module points to some preceding nodes as source nodes; Then, the relation classification module predicts the relation type between the new concept and predicted source nodes. We leave the latter to be determined after iterative inference.
	
\begin{figure}[t]
		\centering
		\includegraphics[scale=0.35]{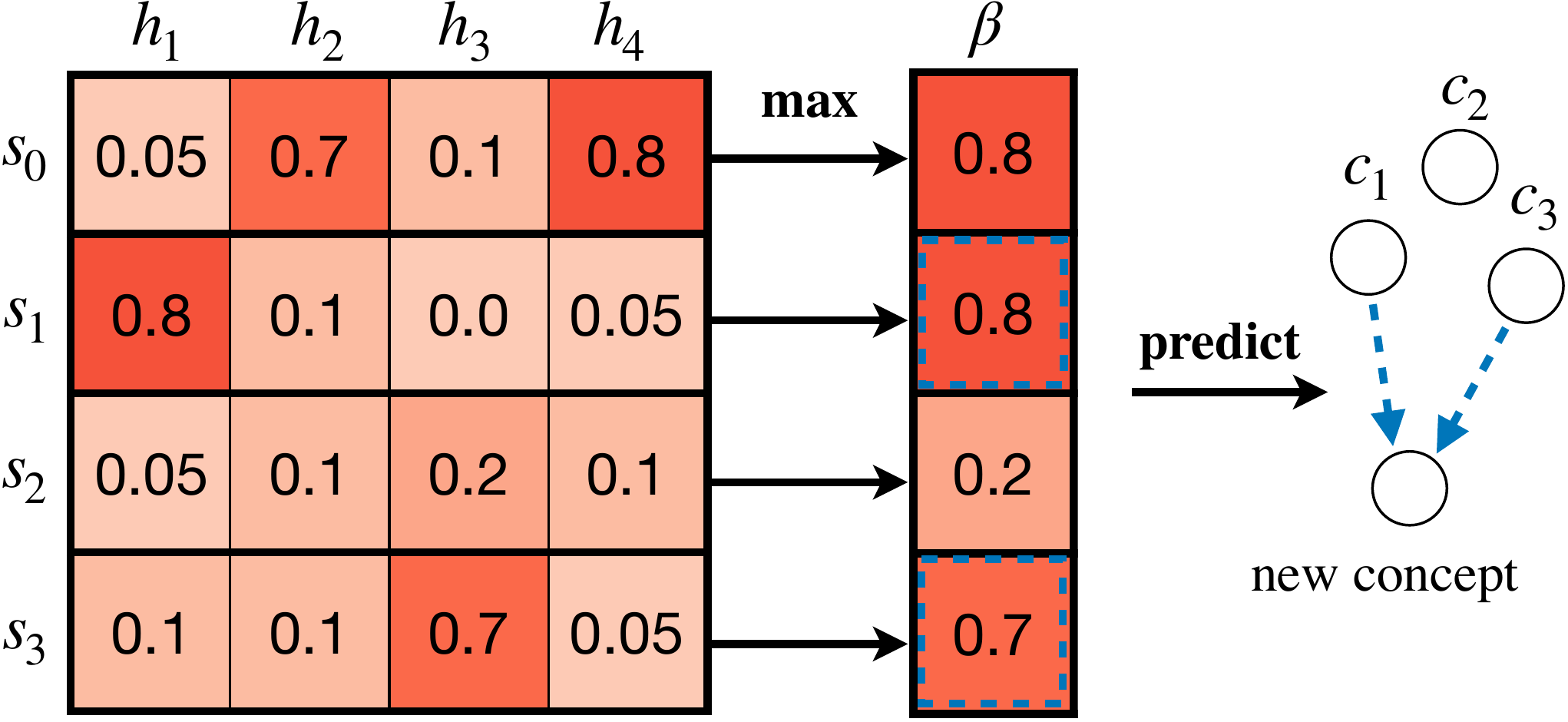}
		\caption{Multi-head attention for relation identification. At left is the attention matrix, where each column corresponds to a unique attention head, and each row corresponds to an existing node.}
		\label{multihead}
	\end{figure}

	AMR is a rooted, directed, and acyclic graph. The reason for AMR being a graph instead of a tree is that it allows reentrancies where a concept participates in multiple semantic relations with different semantic roles. Following \newcite{cai-lam-2019-core}, we use multi-head attention for a more compact parsing procedure where multiple source nodes are simultaneously determined.\footnote{This is different to \newcite{zhang-etal-2019-broad} where an AMR graph is converted into a tree by duplicating nodes that have reentrant relations.} Formally, our relation identification module employs $H$ different attention heads, for each head $h$, we calculate an attention distribution over all existing node (including the dummy node $s_0$):
	\begin{equation}
	\beta^h_t = \text{softmax}(\frac{(W^{Q}_hx_t)^{\text{T}} W^{K}_h s_{0:m}}{\sqrt{d_k}}).
	\nonumber
	\end{equation}
	Then, we take the maximum over different heads as the final edge probabilities:
	\begin{equation}
	\beta_t[i] = \max_{h=1}^H \beta^h_t[i].
	\nonumber
	\end{equation}
	Therefore, different heads may points to different nodes at the same time. Intuitively, each head represents a distinct relation detector for a particular set of relation types. For each attention head, it will point to a source node if certain relations exist between the new node and the existing graph, otherwise it will point to the dummy node. An example with four attention heads and three existing nodes (excluding the dummy node) is illustrated in Figure \ref{multihead}.
	\subsection{Iterative Inference}
	As described above, the concept solver and the relation solver are conceptually two attention mechanisms over the sequence and graph respectively, addressing the concept prediction and relation prediction separately. The key is to pass the decisions between the solvers so that they can examine each other's answer and make harmonious decisions. Specifically, at each spanning step $i$, we start the iterative inference by setting $x_0 = h_0$ and solving $f(G^i, x_0)$. After the $t$-th graph reasoning, we compute the state vector $y_t$, which will be handed over to the concept solver as $g(W, y_t)$, as:
	\begin{equation}
	y_t = \text{FFN}^{(y)}(x_t + (W^{V}h_{1:n}) \alpha_t),
	\nonumber
	\end{equation}
	where $\text{FFN}^{(y)}$ is a feed-forward network and $W^{V}$ projects text memories into a value space. 
	Similarly, after the $t$-th sequence reasoning, we update the state vector from $y_t$ to $x_{t+1}$ as:
	\begin{equation}
	x_{t+1} = \text{FFN}^{(x)}(y_t + \sum_{h=1}^{H}(W^{V}_hs_{0:n})\beta^h_t),
	\nonumber
	\end{equation}
	where $\text{FFN}^{(x)}$ is a feed-forward network and $W^{V}_h$ projects graph memories into a value space for each head $h$. 
	After $N$ steps of iterative inference, i,e., 
	\begin{align*}
	&x_0 \rightarrow f(G^i, x_0) \rightarrow y_1 \rightarrow g(W, y_1) \rightarrow x_1 \rightarrow \cdots\\
	&\rightarrow  f(G^i, x_{N-1}) \rightarrow y_N \rightarrow g(W, y_N) \rightarrow x_{N},
	\end{align*}
	we finally employ a deep biaffine classifier \cite{dozat2016deep} for edge label prediction. The classifier uses a biaffine function to score each label, given the final concept representation $x_N$ and the node vector $s_{1:m}$ as input. The resulted concept, edge, and edge label predictions will added to the new graph $G^{i+1}$ if the concept prediction is not \texttt{EOG}, a special concept that we add for indicating termination. Otherwise, the whole parsing process is terminated and the current graph is returned as final result. The complete parsing process adopting the iterative inference is described in Algorithm \ref{algo}.
	
	\begin{algorithm}[t]
		\caption{AMR Parsing via Graph{\small $\leftrightarrows$}Sequence\\ Iterative Inference }
		\begin{algorithmic}[1]
			\REQUIRE the input sentence $W=(w_1, w_2, \ldots, w_n)$
			\ENSURE the corresponding AMR graph $G$ \\
			\texttt{// compute text memories}
			\STATE $h_0, h_1, \ldots, h_n = $ SequenceEncoder((\texttt{BOS}, $w_1, \ldots, w_n$))\\
			\texttt{// initialize graph}
			\STATE $G^0=($nodes$=\{\texttt{BOG}\}, $edges$=\emptyset)$\\
			\texttt{// start graph expansions}
			\STATE $i=0$
			\WHILE {True}
			\STATE $s_0, \ldots, s_{i}$ = GraphEncoder($G^i$) \\
			\texttt{// the graph memories can be computed *incrementally*}
			\STATE $x_0=h_0$\\
			\texttt{// iterative inference}
			\FOR{$t \gets 1$ to $N$}                    
			\STATE $y_t=f(G^i, x_{t-1})$ \texttt{// Seq.$\rightarrow$Graph}
			\STATE $x_t=g(W, y_t)$ \texttt{// Graph$\rightarrow$Seq.}
			\ENDFOR
			\IF {concept prediction is \texttt{EOG}}
			\STATE \textbf{break}
			\ENDIF
			\STATE update $G^{i+1}$ based on $G^{i}$, $x_N$ and $y_N$
			\STATE $i=i+1$
			\ENDWHILE
			\RETURN $G^i$
		\end{algorithmic}
		\label{algo}
	\end{algorithm}
	\section{Training \& Prediction}
	Our model is trained with the standard maximum likelihood estimate. The optimization objective is to maximize the sum of the decomposed step-wise log-likelihood, where each is the sum of concept, edge, and edge label probabilities. To facilitate training, we create a reference generation order of nodes by running a breadth-first-traversal over target AMR graphs, as it is cognitively appealing \cite[core-semantic-first principle,][]{cai-lam-2019-core} and the effectiveness of pre-order traversal is also empirically verified by \newcite{zhang-etal-2019-amr} in a depth-first setting. For the generation order for sibling nodes, we adopt the uniformly random order and the deterministic order sorted by the relation frequency in a $1:1$ ratio at first then change to the deterministic order only in the final training steps. We empirically find that the deterministic-after-random strategy slightly improves performance.
	
	During testing, our model searches for the best output graph through beam search based on the log-likelihood at each spanning step. The time complexity of our model is $O(k|V|)$, where $k$ is the beam size, and $|V|$ is the number of nodes.
	\section{Experiments}
	\subsection{Experimental Setup}
	\paragraph{Datasets} Our evaluation is conducted on two AMR public releases: AMR 2.0 (LDC0217T10) and AMR 1.0 (LDC2014T12). AMR 2.0 is the latest and largest AMR sembank that was extensively used in recent works. AMR 1.0 shares the same development and test set with AMR, while the size of its training set is only about one-third of AMR 2.0, making it a good testbed to evaluate our model's sensitivity for data size.\footnote{There are a few annotation revisions from AMR 1.0 to AMR 2.0.}
	\paragraph{Implementation Details } We use Stanford CoreNLP \cite{manning2014stanford} for tokenization, lemmatization, part-of-speech, and named entity tagging. The hyper-parameters of our models are chosen on the development set of AMR 2.0. Without explicit specification, we perform $N=4$ steps of iterative inference. Other hyper-parameter settings can be found in the Appendix. Our models are trained using ADAM \cite{kingma2014adam} for up to 60K steps (first 50K with the random sibling order and last 10K with deterministic order), with early stopping based on development set performance. We fix BERT parameters similar to \newcite{zhang-etal-2019-amr,zhang-etal-2019-broad} due to the GPU memory limit. During testing, we use a beam size of 8 for the highest-scored graph approximation.\footnote{Our code is released at \url{https://github.com/jcyk/AMR-gs}.}
	\paragraph{AMR Pre- and Post-processing} We remove senses as done in \newcite{lyu2018amr,zhang-etal-2019-amr,zhang-etal-2019-broad} and simply assign the most frequent sense for nodes in post-processing. Notably, most existing methods including the state-the-of-art parsers \cite[][inter alia]{zhang-etal-2019-amr,zhang-etal-2019-broad,lyu2018amr,guo2018better} often rely on heavy graph re-categorization for reducing the complexity and sparsity of the original AMR graphs. For graph re-categorization, specific subgraphs of AMR are grouped together and assigned to a single node with a new compound category, which usually involves non-trivial expert-level manual efforts for hand-crafting rules. We follow the exactly same pre- and post-processing steps of those of \newcite{zhang-etal-2019-amr,zhang-etal-2019-broad} for graph re-categorization. More details can be found in the Appendix.
	\paragraph{Ablated Models}
	As pointed out by \newcite{cai-lam-2019-core}, the precise set of graph re-categorization rules differs among different works, making it difficult to distinguish the performance improvement from model optimization and carefully designed rules. In addition, only recent works \cite{zhang-etal-2019-amr,zhang-etal-2019-broad,lindemann-etal-2019-compositional,naseem-etal-2019-rewarding} have started to utilize the large-scale pre-trained language model, BERT \cite{devlin-etal-2019-bert,Wolf2019HuggingFacesTS}. Therefore, we also include ablated models for addressing two questions: (1) How dependent is our model on performance from hand-crafted graph re-categorization rules? (2) How much does BERT help? We accordingly implement three ablated models by removing either one of them or removing both. The ablation study not only reveals the individual effect of two model components but also helps facilitate fair comparisons with prior works.
	\subsection{Experimental Results}
	\begin{table*}[t]
		\small
		\centering
		\resizebox{2.1\columnwidth}{!}{
			\begin{tabular}{c|c|c|c||c|c|c|c|c|c|c|c}
				\hline
				\multirow{2}{*}{Model}&\multirow{2}{*}{\scriptsize{G. R.}}&\multirow{2}{*}{\scriptsize{BERT}}&\multirow{2}{*}{\scriptsize{\textsc{Smatch}}}&\multicolumn{8}{c}{fine-grained evaluation}\\
				\cline{5-12}
				& &  & & \scriptsize{Unlabeled} & \scriptsize{No WSD} & \scriptsize{Concept} & \scriptsize{SRL} &  \scriptsize{Reent.} &\scriptsize{Neg.} & \scriptsize{NER}& \scriptsize{Wiki}\\
				\hline
				\newcite{van2017neural} &\texttimes& \texttimes& 71.0 & 74 & 72 & 82 & 66 & 52 & 62 & 79 & 65\\ 
				\newcite{groschwitz2018amr} &$\checkmark$& \texttimes& 71.0&74&72 & 84 & 64 & 49 & 57&78&71\\
				\newcite{lyu2018amr} &$\checkmark$& \texttimes& 74.4 &77.1 & 75.5 & 85.9 & 69.8 & 52.3 & 58.4 & 86.0 & 75.7\\
				\newcite{cai-lam-2019-core} &\texttimes& \texttimes& 73.2 & 77.0& 74.2&84.4&66.7&55.3&62.9&82.0&73.2\\
				\newcite{lindemann-etal-2019-compositional} &$\checkmark$& $\checkmark$&75.3 &-&-&-&-&-&-&-&-\\
				\newcite{naseem-etal-2019-rewarding} &$\checkmark$& $\checkmark$& 75.5 & 80 & 76 & 86 & 72 & 56 & 67 & 83& 80\\
				\newcite{zhang-etal-2019-amr}    &$\checkmark$& \texttimes&74.6 &-&-&-&-&-&-&-&-\\
				\newcite{zhang-etal-2019-amr} &$\checkmark$& $\checkmark$& 76.3&79.0&76.8&84.8&69.7&60.0&75.2&77.9&85.8\\
				\newcite{zhang-etal-2019-broad} &$\checkmark$& $\checkmark$& 77.0 & 80& 78& 86&71&61&77&79&86\\
				\hline
				\multirow{4}{*}{Ours} &\texttimes& \texttimes& 74.5&77.8&75.1&85.9&68.5&57.7&65.0&82.9&81.1\\
				&$\checkmark$& \texttimes& 77.3&80.1&77.9&86.4&69.4&58.5&75.6&78.4&86.1\\
				&\texttimes &$\checkmark$& 78.7&81.5&79.2&88.1&\textbf{74.5}&63.8&66.1&\textbf{87.1}&81.3\\
				&$\checkmark$& $\checkmark$& \textbf{80.2}&\textbf{82.8}&\textbf{80.8}&\textbf{88.1}&74.2&\textbf{64.6}&\textbf{78.9}&81.1&\textbf{86.3}\\
				\hline
		\end{tabular}}
		\caption{\textsc{Smatch} scores (\%) (left) and fine-grained evaluations (\%) (right) on the test set of AMR 2.0. G. R./BERT indicates whether or not the results use Graph Re-categorization/BERT respectively.}
		\label{amr2}
	\end{table*}
	\begin{table}[t]
		\centering
		\small
		\begin{tabular}{c|c|c|c}
			\hline
			Model & \scriptsize{G. R.}& \scriptsize{BERT} & \scriptsize{\textsc{Smatch}} \\
			\hline
			\newcite{flanigan2016cmu}&\texttimes& \texttimes& 66.0 \\
			\newcite{pust2015parsing} &\texttimes& \texttimes& 67.1 \\
			\newcite{wang2017getting} &$\checkmark$& \texttimes& 68.1\\
			\newcite{guo2018better} &$\checkmark$& \texttimes& 68.3 \\
			\newcite{zhang-etal-2019-amr} &$\checkmark$& $\checkmark$& 70.2\\
			\newcite{zhang-etal-2019-broad} &$\checkmark$& $\checkmark$& 71.3 \\
			\hline
			\multirow{4}{*}{Ours} &\texttimes& \texttimes& 68.8\\
			&$\checkmark$& \texttimes& 71.2\\
			&\texttimes &$\checkmark$& 74.0\\
			&$\checkmark$& $\checkmark$&\textbf{75.4}\\
			\hline
		\end{tabular}
		\caption{\textsc{Smatch} scores on the test set of AMR 1.0.}
		\label{amr1}
	\end{table}
	\paragraph{Main Results} The performance of AMR parsing is conventionally evaluated by \textsc{Smatch} (F1) metric \cite{cai2013smatch}. The left block of Table \ref{amr2} shows the \textsc{Smatch} scores on the AMR 2.0 test set of our models against the previous best approaches and recent competitors. On AMR 2.0, we outperform the latest push from \newcite{zhang-etal-2019-broad} by 3.2\% and, for the first time, obtain a parser with over 80\% \textsc{Smatch} score. Note that even without BERT, our model still outperforms the previous state-of-the-art approaches using BERT \cite{zhang-etal-2019-broad,zhang-etal-2019-amr} with 77.3\%. This is particularly remarkable since running BERT is computationally expensive. As shown in Table \ref{amr1}, on AMR 1.0 where the training instances are only around 10K, we improve the best-reported results by 4.1\% and reach at 75.4\%, which is already higher than most models trained on AMR 2.0. The even more substantial performance gain on the smaller dataset suggests that our method is both effective and data-efficient. Besides, again, our model without BERT already surpasses previous state-of-the-art results using BERT. For ablated models, it can be observed that our models yield the best results in all settings if there are any competitors, indicating BERT and graph re-categorization are not the exclusive key for our superior performance.
	\paragraph{Fine-grained Results} In order to investigate how our parser performs on individual sub-tasks, we also use the fine-grained evaluation tool \cite{damonte2016incremental} and compare to systems which reported these scores.\footnote{We only list the results on AMR 2.0 since there are few results on AMR 1.0 to compare.} As shown in the right block of Table \ref{amr2}, our best model obtains the highest scores on almost all sub-tasks. The improvements in all sub-tasks are consistent and uniform (around 2\%$\sim$3\%) compared to the previous state-of-the-art performance \cite{zhang-etal-2019-broad}, partly confirming that our model boosts performance via consolidated and harmonious decisions rather than fixing particular phenomena. By our ablation study, it is worth noting that the NER scores are much lower when using graph re-categorization. This is because the rule-based system for NER in graph re-categorization does not generalize well to unseen entities, which suggest a potential improvement by adapting better NER taggers.
	\subsection{More Analysis}
		\begin{figure}[t]
		\centering
		\includegraphics[scale=0.18]{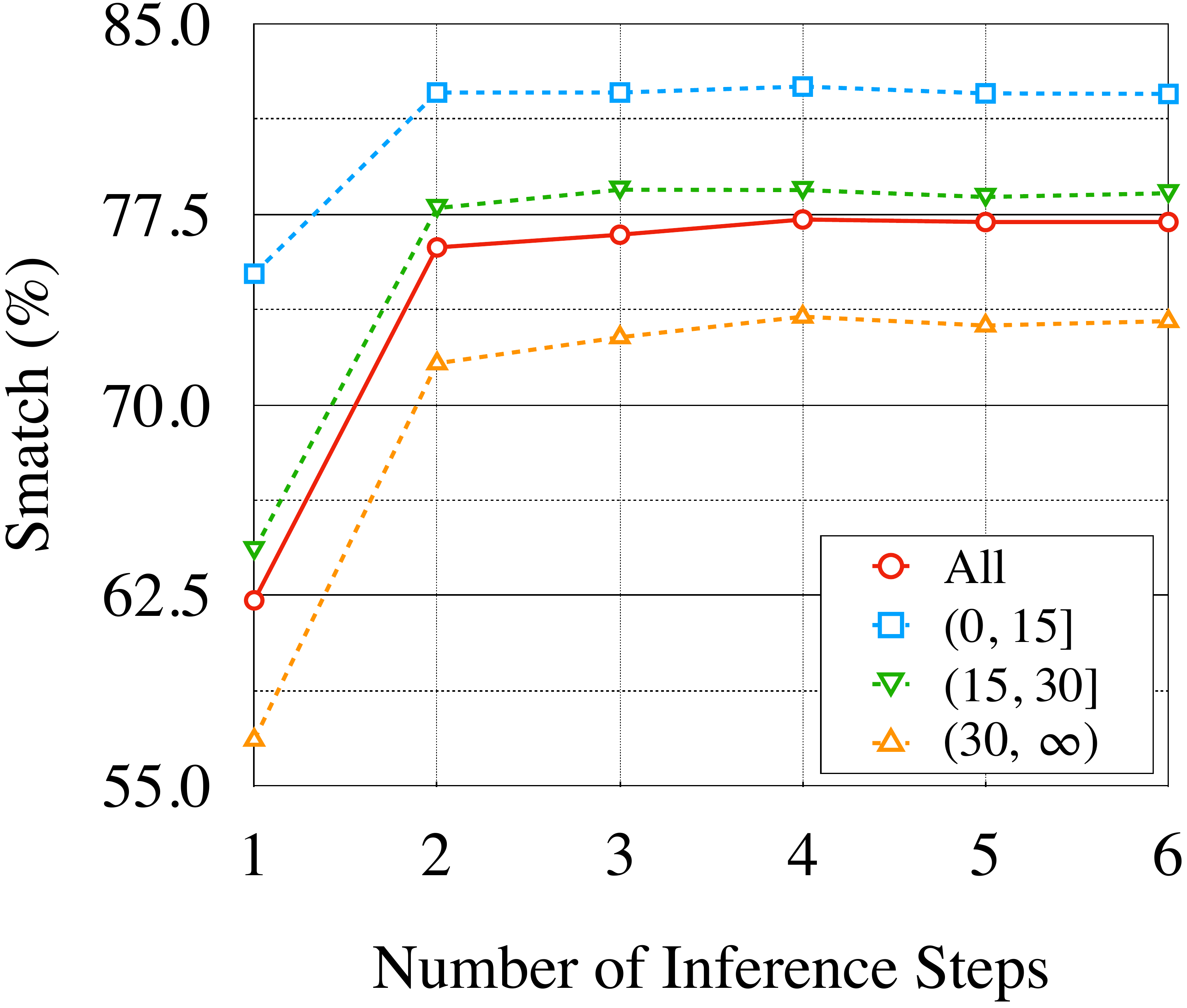}
		\caption{\textsc{Smatch} scores with different numbers of inference steps. Sentences are grouped by length.}
		\label{step}
	\end{figure}
	\begin{figure*}[t]
		\centering
		\includegraphics[scale=0.43]{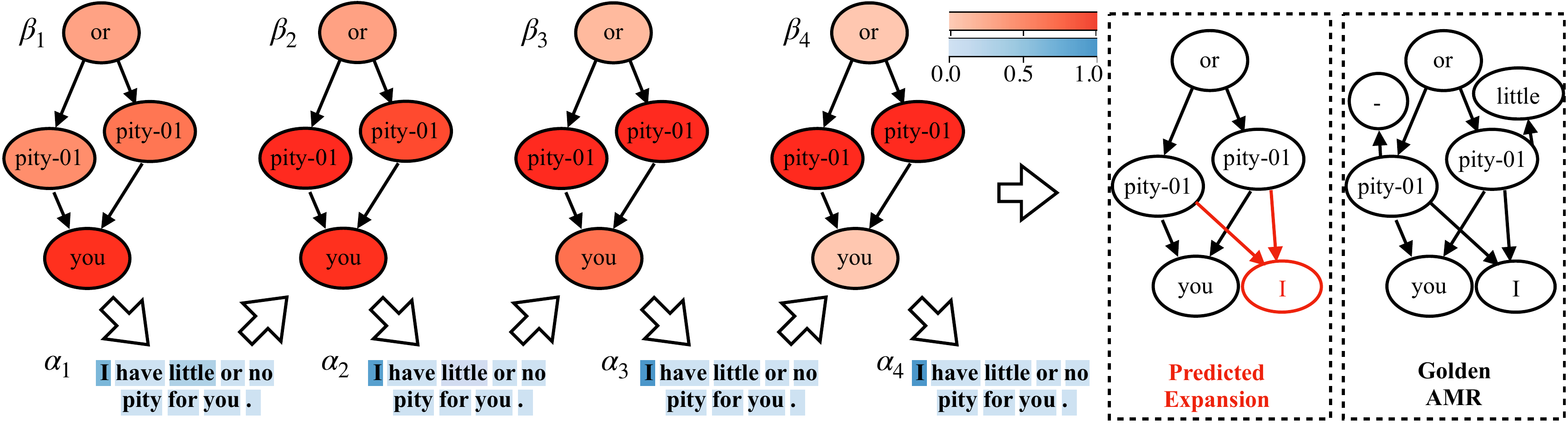}
		\caption{Case study (viewed in color). Color shading intensity represents the value of the attention score.}
		\label{case}
	\end{figure*}
	\paragraph{Effect of Iterative Inference}
	We then turn to study the effect of our key idea, namely, the iterative inference design. To this end, we run a set of experiments with different values of the number of the inference steps $N$. The results on AMR 2.0 are shown in Figure \ref{step} (solid line). As seen, the performance generally goes up when the number of inference steps increases. The difference is most noticeable between 1 (\textit{no iterative} reasoning is performed) and 2, while later improvements gradually diminish. One important point here is that the model size in terms of the number of parameters is constant regardless of the number of inference steps, making it different from general over-parameterized problems.
	
\begin{figure}[t]
		\centering
		\includegraphics[scale=0.18]{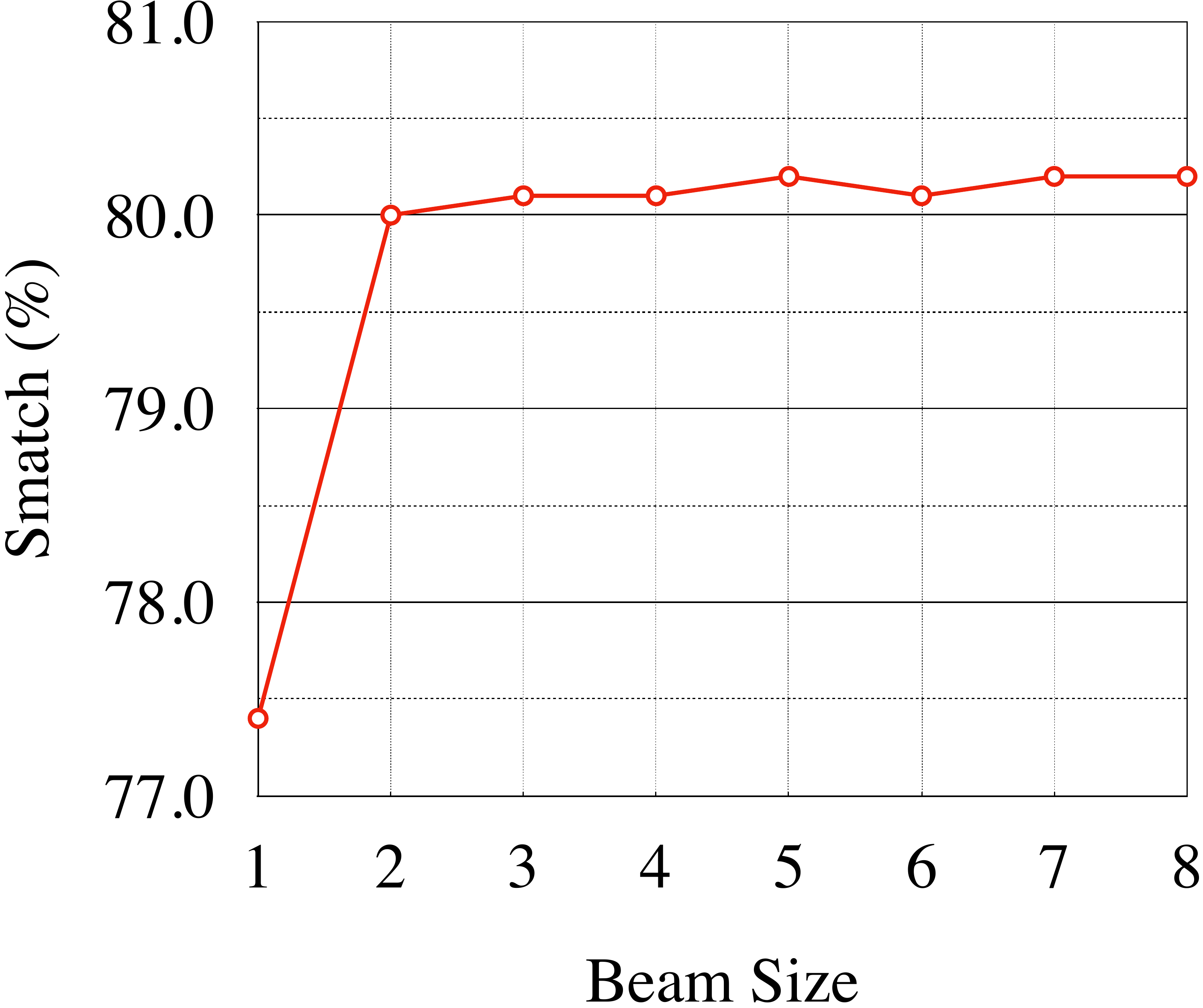}
		\caption{\textsc{Smatch} scores with different beam sizes.}
		\label{beam}
	\end{figure}

	For a closer study on the effect of the inference steps with respect to the lengths of input sentences, we group sentences into three classes by length and also show the individual results in Figure \ref{step} (dashed lines). As seen, the iterative inference helps more for longer sentences, which confirms our intuition that longer and more complex input needs more reasoning. Another interesting observation is that the performance on shorter sentences reaches the peaks earlier. This observation suggests that the number of inference steps can be adjusted according to the input sentence, which we leave as future work.
	\paragraph{Effect of Beam Size}
	We are also interested in the effect of beam size during testing. Ideally, if a model is able to make accurate predictions in the first place, it should rely less on the search algorithm. We vary the beam size and plot the curve in Figure \ref{beam}. The results show that the performance generally gets better with larger beam sizes. However, a small beam size of 2 already gets the most of the credits, which suggests that our model is robust enough for time-stressing environments.
	\paragraph{Visualization}
	We visualize the iterative reasoning process with a case study in Figure \ref{case}. We illustrate the values of $\alpha_t, \beta_t$ as the iterative inference progresses. As seen, the parser makes mistakes in the first step, but gradually corrects its decisions and finally makes the right predictions. Later reasoning steps typically provide a sharper attention distribution than earlier steps, narrowing down the most likely answer with more confidence.
	\paragraph{Speed}
	We also report the parsing speed of our non-optimized code: With BERT, the parsing speed of our system is about 300 tokens/s, while without BERT, it is about 330 tokens/s on a single Nvidia P4 GPU. The absolute speed depends on various implementation choices and hardware performance. In theory, the time complexity of our parsing algorithm is $O(kbn)$, where $k$ is the number of iterative steps, $b$ is beam size, and $n$ is the graph size (number of nodes) respectively. It is important to note that our algorithm is linear in the graph size.
	\section{Conclusion}
	We presented the dual graph-sequence iterative inference method for AMR Parsing. Our method constructs an AMR graph incrementally in a node-by-node fashion. Each spanning step is explicitly characterized as answering two questions: which parts of the sequence to abstract, and where in the graph to construct. We leverage the mutual causalities between the two and design an iterative inference algorithm. Our model significantly advances the state-of-the-art results on two AMR corpora. An interesting future work is to make the number of inference steps adaptive to input sentences. Also, the idea proposed in this paper may be applied to a broad range of structured prediction tasks (not only restricted to other semantic parsing tasks) where the complex output space can be divided into two interdependent parts with a similar iterative inference process to achieve harmonious predictions and better performance.
	\bibliography{acl2020}
	\bibliographystyle{acl_natbib}

\appendix
\newpage
\section{Hyper-parameter Settings}
Table \ref{setting} lists the hyper-parameters used in our full models. Char-level CNNs and Transformer layers in the sentence encoder and the graph encoder share the same hyper-parameter settings. The BERT model \cite{devlin-etal-2019-bert} we used is the Huggingface’s implementation \cite{Wolf2019HuggingFacesTS} (bert-base-cased). To mitigate overfitting, we apply dropout \cite{srivastava2014dropout} with the drop rate $0.2$ between different layers. We randomly mask (replacing inputs with a special UNK token) the input lemmas, POS tags, and NER tags with a rate of $0.33$. Parameter optimization is performed with the ADAM optimizer \cite{kingma2014adam} with $\beta_1=0.9$ and $\beta_2=0.999$. The learning rate schedule is similar to that in \newcite{vaswani2017attention}, with warm-up steps being set to 2K. We use early stopping on the development set for choosing the best model.
\section{AMR Pre- and Post-processing}
We follow exactly the same pre- and post-processing steps of those of \newcite{zhang-etal-2019-amr,zhang-etal-2019-broad} for graph re-categorization. In preprocessing, we anonymize entities, remove wiki links and polarity attributes, and convert the resultant AMR graphs into a compact format by compressing certain subgraphs. In post-processing, we recover the original AMR format from the compact format, restore Wikipedia links using the DBpedia Spotlight API \cite{daiber2013improving}, add polarity attributes based on rules observed from the training data. More details can be found in \newcite{zhang-etal-2019-amr}.
\begin{table}[t]
	\centering
	\begin{tabular}{lr}
		\hline
		\multicolumn{2}{l}{\textbf{Embeddings}} \\
		lemma & 300 \\
		POS tag& 32 \\
		NER tag & 16 \\
		concept & 300 \\
		char & 32 \\
		\hline
		\multicolumn{2}{l}{\textbf{Char-level CNN}} \\
		\#filters & 256 \\
		ngram filter size & [3] \\
		output size & 128 \\
		\hline
		\multicolumn{2}{l}{\textbf{Sentence Encoder}} \\
		\#transformer layers & 4 \\
		\hline
		\multicolumn{2}{l}{\textbf{Graph Encoder}} \\
		\#transformer layers & 2 \\
		\hline
		\multicolumn{2}{l}{\textbf{Transformer Layer}} \\
		\#heads & 8 \\
		hidden size & 512 \\
		feed-forward hidden size & 1024 \\
		\hline
		\multicolumn{2}{l}{\textbf{Concept Solver}} \\
		feed-forward hidden size & 1024 \\
		\hline
		\multicolumn{2}{l}{\textbf{Relation Solver}} \\
		\#heads & 8 \\
		feed-forward hidden size & 1024 \\
		\hline
		\multicolumn{2}{l}{\textbf{Deep biaffine classifier}} \\
		hidden size & 100 \\
		\hline
	\end{tabular}
	\caption{Hyper-parameters settings. }
	\label{setting}
\end{table}
\end{document}